%% file: main.tex
\documentclass[10pt]{article} %
\usepackage[preprint]{tmlr}

\input{math_commands.tex}

\usepackage{hyperref}
\usepackage{url}
\usepackage{xspace}
\usepackage{graphicx} 
\usepackage{latexsym}
\usepackage{tabularx}
\usepackage[utf8]{inputenc}
\usepackage{package}

\usepackage{comment}

\title{PixelWorld: How Far Are We from Perceiving Everything as Pixels?\thanks{Accepted by TMLR.}}

\author{
  Zhiheng Lyu$^{1,2}$\quad
  Xueguang Ma$^{1}$\quad
  Wenhu Chen$^{1,2}$ \\
  $^1$University of Waterloo \quad
  $^2$Vector Institute, Toronto \\
  \texttt{\{z63lyu,x93ma,wenhuchen\}@uwaterloo.ca}
}

\newcommand{\method}{\textsc{PEAP}\xspace}

\def\month{MM}  %
\def\year{YYYY} %
\def\openreview{\url{https://openreview.net/forum?id=XXXX}} %

\begin{document}

\maketitle

\begin{abstract}
Recent agentic language models increasingly need to interact with real-world environments that contain tightly intertwined visual and textual information, often through raw camera pixels rather than separately processed images and tokenized text. This shift highlights the need for a \emph{unified perception} paradigm. To investigate this idea, we explore \textbf{Perceive Everything as Pixels} (\method) and introduce \textsc{PixelWorld}, a benchmark that renders natural-language, tabular, mathematical, and diagrammatic inputs into a shared pixel space. Experiments across multiple benchmarks show that \method\ achieves comparable performance to token-based approaches on semantic understanding tasks, suggesting that vision transformers can partially capture global textual semantics without explicit tokenization. In contrast, reasoning-intensive tasks such as mathematics and code show notable performance degradation, although Chain-of-Thought prompting helps mitigate this gap by compensating for missing symbolic structure. We further find that when visual and textual information are closely integrated, representing everything as pixels simplifies preprocessing and avoids cross-modal misalignment. \textsc{PixelWorld} thus provides a systematic and practical framework for evaluating unified vision--language models and facilitates further exploration of pixel-based multimodal learning.
\end{abstract}

\section{Introduction}
In recent years, large vision--language models (L-VLMs)~\citep{Qwen2VL,openai_gpt4o,gemini} have achieved remarkable progress across diverse real-world tasks. However, these models still rely on distinct processing pipelines for different modalities---treating images as pixels and text as discrete tokens. Such disjoint tokenization introduces a \emph{representation mismatch} between vision and language, which hampers unified multimodal understanding and complicates system design. As recent works~\citep{seeact,VisualWebArena,tellex2020robots,PALME} push toward \textbf{agentic systems} capable of perceiving and acting in complex environments---from physical navigation~\citep{elnoor2024robot} and travel booking~\citep{chen2024travelagent} to code repair on GitHub~\citep{yang2024swe}---this mismatch becomes increasingly consequential. In such intertwined visual--textual settings, maintaining separate tokenization and perception modules not only incurs high preprocessing overhead~\citep{OSWorld,VisualWebArena} but also leads to information loss and layout inconsistencies~\citep{dagangetting,chai2024tokenization}, ultimately limiting the scalability and robustness of multimodal agents.

To address these limitations, we explore a unified perception approach: \textbf{Perceive Everything as Pixels} (\method). Building on earlier efforts that considered pixel-based representations~\citep{singh2024captioningtaskspecificpromptingimproved,zhang2024improvevisionlanguagemodel}, we systematically examine how representing both text and visual inputs uniformly in pixel space affects model behavior. In this paradigm, a vision--language model (VLM) jointly models multimodal inputs without requiring separate tokenization or modality-specific encoders. To better understand the benefits and challenges of this approach, we introduce \textsc{PixelWorld}, a comprehensive benchmark suite designed to evaluate how well existing VLMs perform under the \method setting. 

\begin{figure*}[h]
    \centering
    \includegraphics[width=0.9\textwidth]{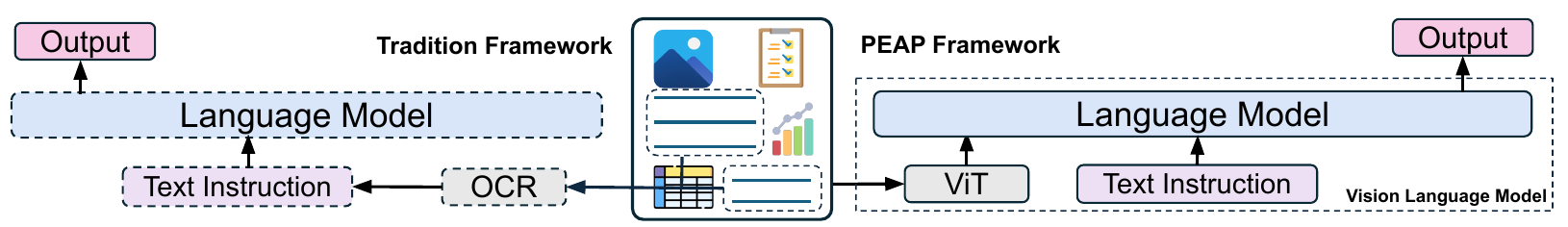}
    \caption{\textbf{Overview of the \method\ Framework.}
    \method\ (\emph{Perceive Everything as Pixels}) unifies text, structural, and visual inputs into a single pixel space, where a Vision Transformer (ViT) encodes the pixels and a language decoder performs reasoning. 
    Both components are enclosed within the dashed box to indicate that they jointly form a vision–language model (VLM). 
    By eliminating modality-specific preprocessing such as OCR and tokenization, \method\ better aligns with human perception and reduces cross-modal engineering overhead.}
    \label{fig:peap_framework}
\end{figure*}

\begin{figure*}[h]
    \centering
    \includegraphics[width=0.9\textwidth]{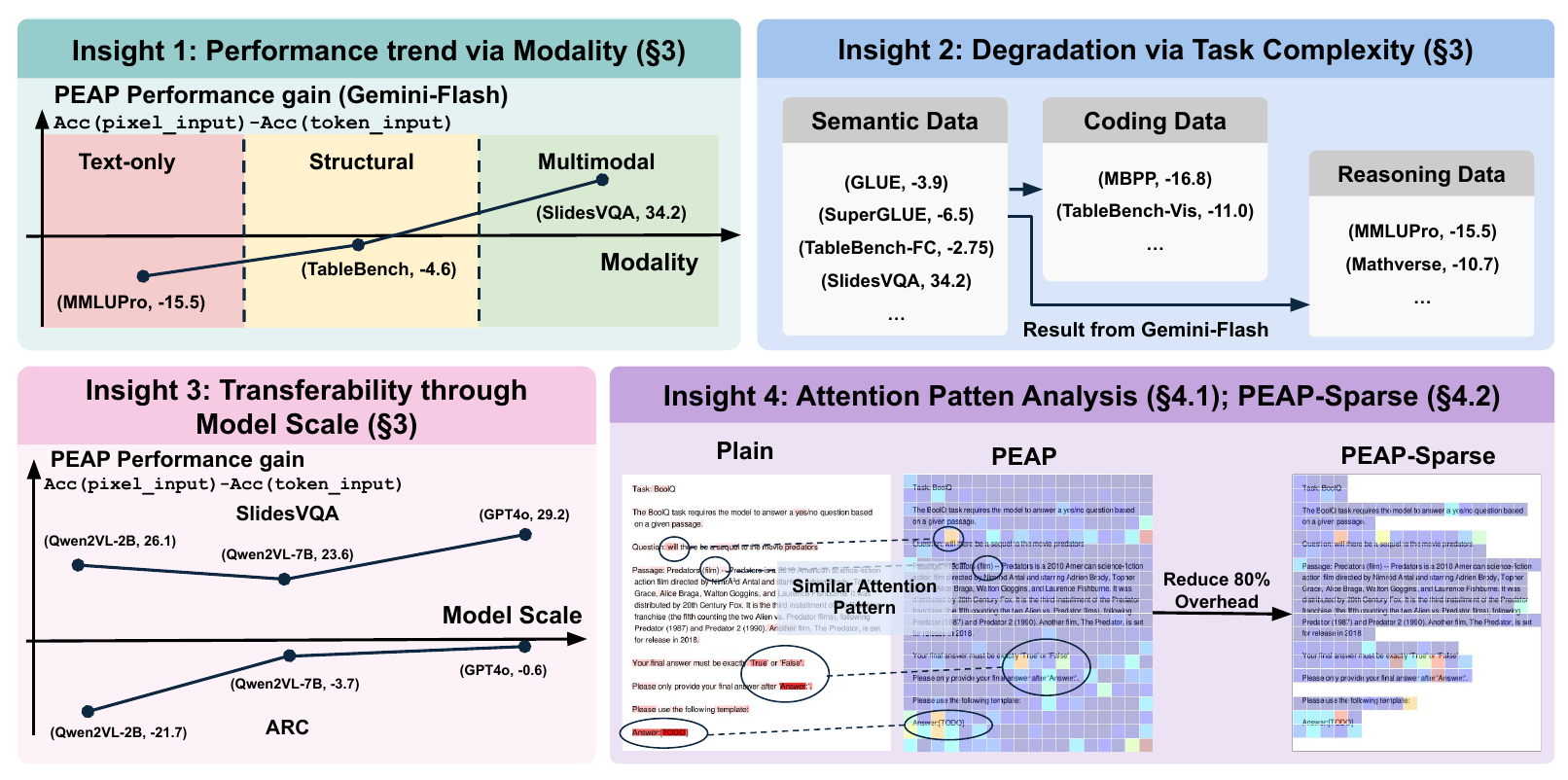}
    \caption{\textbf{Key Findings on the \textsc{PixelWorld} Benchmark.}
    Evaluated across text-only, structural, and multimodal settings (§\ref{sec:data}, §\ref{sec:experiment}), \method\ shows four major insights: 
    (1)~\emph{Modality Trend:} consistent gains on layout-heavy and multimodal tasks such as websites, slides, and documents; 
    (2)~\emph{Task Complexity:} performance degradation on reasoning- and code-centric benchmarks (see §\ref{sec:experiment_text}–§\ref{sec:experiment_structure}); 
    (3)~\emph{Transferability by Scale:} larger VLMs (e.g., GPT-4o, Gemini-Flash) exhibit smaller pixel–token gaps; and 
    (4)~\emph{Attention and Efficiency:} text and image inputs show similar global attention patterns, while the proposed \method-Fast reduces up to 80\% of computation overhead (§\ref{sec:dis_2}).}
    \label{fig:peap}
\end{figure*}
In \textsc{PixelWorld}, we select 10 widely used benchmarks covering a diverse range of modalities and task scenarios. For each dataset, we construct both traditional token-based and pixel-based (\method) input formats using image synthesis and OCR techniques (see Table~\ref{tab:dataset_overview}). We then evaluate vision--language models of varying scales, from Qwen2VL-2B to GPT-4o. Cross-modal evaluation in Section~\ref{sec:experiment} reveals three overarching insights:  
\textbf{Insight 1}: In intrinsically multimodal settings such as website rendering, slide comprehension, and document understanding, \method consistently mitigates OCR noise and yields stronger performance;  
\textbf{Insight 2}: For reasoning-intensive tasks such as mathematics and code, pixelization leads to noticeable accuracy drops, though the gap narrows as model capacity increases---suggesting that scale plays a key role in enabling cross-modal transfer; and  
\textbf{Insight 3}: Larger models exhibit more robust instruction-following and long-context reasoning across modalities, whereas smaller models struggle, highlighting the importance of scale-aware training under the pixel-based paradigm.

To further interpret these findings, we conduct three complementary analyses.  
{(1) Representation analysis:} We visualize the attention patterns of Qwen2VL-7B and observe broadly similar global structures between token- and pixel-based inputs, suggesting that certain aspects of language modeling behavior may transfer into the visual space, though not implying full equivalence.  
{(2) Efficiency optimization:} We measure inference latency and find that while \method\ increases computational cost due to larger input resolution, our proposed \method-Fast algorithm effectively prunes blank patches, achieving up to 80\% speedup with negligible accuracy degradation.  
{(3) Prompt sensitivity:} We study prompting strategies and find that Chain-of-Thought (CoT) reasoning yields more consistent gains under the pixel-based representation compared to standard direct prompting, indicating potential synergies between reasoning supervision and visual encoding.

In summary, our contributions are as follows:
\begin{enumerate}[
    itemsep=0pt,
    parsep=0pt,
    topsep=0pt,
    leftmargin=1em
]
\item \textbf{\textsc{PixelWorld} Benchmark}: We present a unified benchmark that transforms text, structural, and multimodal datasets into pixel space, offering a direct and reproducible framework to evaluate the trade-offs between pixel- and token-based modeling. The benchmark and code are publicly released to facilitate standardized comparison and future research on multimodal perception.
\item \textbf{Task–scale insights}: Through large-scale evaluation, we show that \method\ improves layout-heavy or intrinsically multimodal tasks (e.g., website and document understanding) while reducing accuracy on reasoning- or code-centric tasks. The performance gap consistently narrows with model scale, underscoring the role of capacity in enabling cross-modal transfer.
\item \textbf{Efficiency \& interpretability}: We propose \method-Fast, an inference-time pruning strategy that removes blank pixel patches, achieving up to a \(3\times\) latency reduction with minimal loss in accuracy. Attention visualizations reveal partially shared global structures across modalities, providing an interpretable perspective on how visual encoders approximate token-level reasoning behavior.
\end{enumerate}

\begin{table*}[t]
    \centering 
    \footnotesize
    \renewcommand{\arraystretch}{1.2} %
    \setlength{\tabcolsep}{6pt} %
    \begin{tabular}{l c l c c}
    \toprule
    \textbf{Dataset Name} & \textbf{Size} & \textbf{Task} & \textbf{Modality Transfer} & \textbf{Split}\\
    \midrule
    \multicolumn{5}{c}{\textbf{Text-only}} \\
    \midrule
    GLUE \citep{wang2018glue} & 59,879 & Natural language understanding & Synthesis & test \\
    SuperGLUE \citep{sarlin2020superglue} & 19,294 & Natural language understanding & Synthesis & test \\
    MMLU-Pro \citep{wang2024mmlu} & 12,032 & Domain knowledge and reasoning & Synthesis & test \\
    ARC \citep{allenai:arc} & 3,548 & Science question answering & Synthesis & test \\
    GSM8K \citep{cobbe2021gsm8k} & 1,319 & Math problem solving & Synthesis & test \\
    MBPP \citep{austin2021program} & 757 & Programming tasks & Synthesis & test \\
    \midrule
    \multicolumn{5}{c}{\textbf{Structured}} \\
    \midrule
    TableBench \citep{wu2024tablebench} & 888 & Table data understanding and analysis & Synthesis & test \\
    \midrule
    \multicolumn{5}{c}{\textbf{Multimodal}} \\
    \midrule
    MathVerse \citep{zhang2025mathverse} & 788 & Math and visual reasoning & Natural & test \\
    MMMU-Pro \citep{yue2024mmmuprorobustmultidisciplinemultimodal} & 1,730 & Multimodal reasoning & Synthesis & test \\
    SlidesVQA~\citep{tanaka2023slidevqadatasetdocumentvisual} & 2,136 & Multimodal question answering & OCR & test \\
    Wiki-SS~\citep{ma-etal-2024-unifying} & 3,000 & Multimodal retrieval question answering & OCR & train \\
    \bottomrule
    \end{tabular}
    \caption{Overview of datasets categorized by modality, usage, size, and split. Modality Transfer means the method to adopt the dataset into counterpart modality. For OCR, we adopt the result from the origin datasets. For WikiSS-QA, since the positive document of the test set is not released, we subsample 3,000 training data points randomly to evaluate.}
    \label{tab:dataset_overview}
\end{table*}

\section{Datasets}
\label{sec:data}
Several representative datasets covering different skill domains are selected, as shown in Table \ref{tab:dataset_overview}. We primarily utilize the prompts provided by the datasets. If no prompts are available, we apply a default prompt. By default, we employ Direct Prompting; however, for more complex and mathematical datasets such as MBPP~\citep{austin2021program}, MMLU-Pro~\citep{wang2024mmlu}, and MathVerse~\citep{zhang2025mathverse}, we adopt Chain-of-Thought (CoT) prompting to enhance performance. All evaluations are conducted in a zero-shot manner to mitigate potential performance degradation caused by the sensitivity of instruction-tuned large models to few-shot prompting.

To evaluate both Token-based and Pixel-based methods, we require paired Text-input and Image-input prompts. We adopted modality transfer strategies to reduce reliance on the information modality provided by existing datasets, as detailed in Table \ref{tab:dataset_overview}. For datasets categorized as \textit{Text-Only} and \textit{Structured}, all data is originally in plain text format, necessitating image synthesis prior to evaluation. For \textit{Multimodal} datasets, textual content embedded in images is extracted using OCR, or the textual components provided by the original datasets are directly utilized for evaluation. Notably, the MathVerse dataset~\citep{zhang2025mathverse} inherently includes a Text-Only modality, offering detailed textual descriptions of image-based information.

\paragraph{Image Data Synthesis} For text-only and structured datasets, we developed an image data synthesis pipeline to generate diverse image inputs for evaluation. Image widths were adaptively adjusted between 512 and 1024 pixels based on text length, with a fixed height of 256 pixels. Font sizes ranged from 15 to 25 points, and padding varied from 5 to 30 pixels. To enhance robustness, we applied various types of noise, including radial, horizontal, vertical, and Multi-Gaussian noise, as well as high-frequency Gaussian noise to simulate distortions commonly introduced by real-world cameras. For structured datasets, such as tables, data was rendered as images using the Python package \texttt{dataframe\_image}. Example inputs from different tasks are provided in Appendix \ref{sec:example_input}.

\section{Experiments}
In this section, we will detail our baseline, metrics and models. The experimental results will be organized by `Text Input', `Structued Input' and `Multimodal Input'. 

\begin{figure*}[t]
    \centering
    \includegraphics[width=\textwidth]{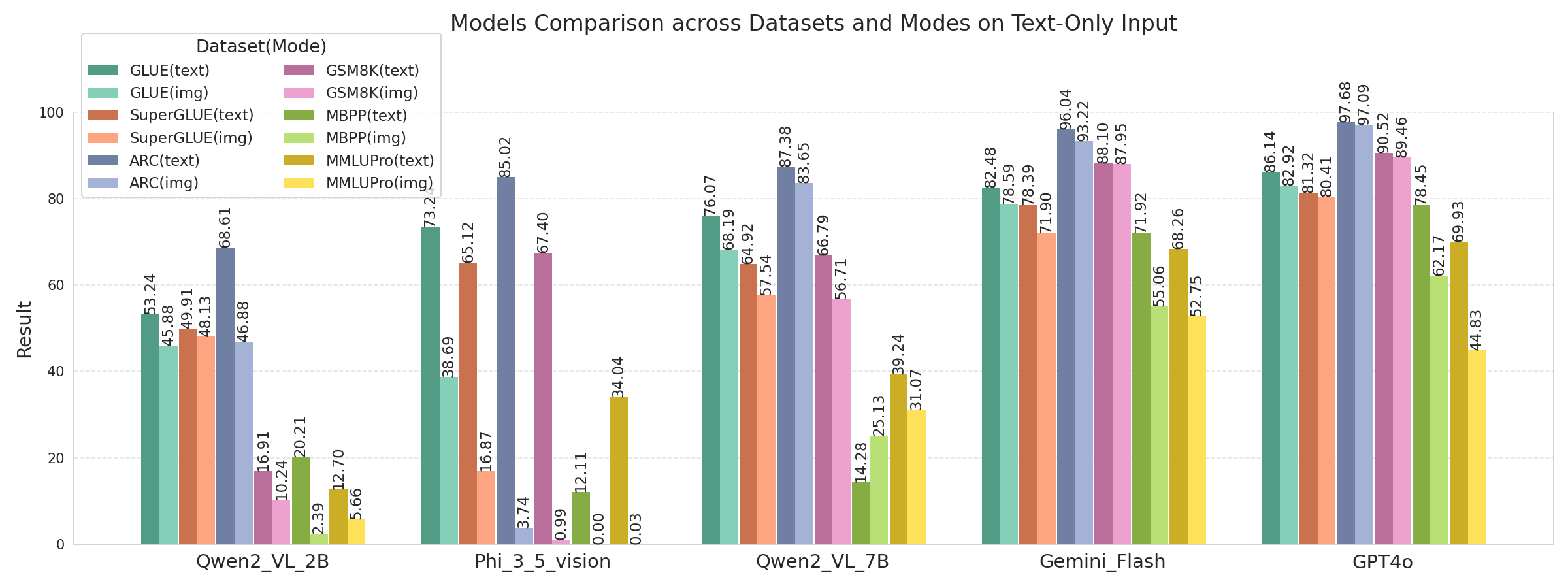} 
    \caption{The performance of \textbf{text-only} datasets. The comparison is made between text input and synthesized image input. Most models demonstrate comparable performance on language understanding datasets such as SuperGLUE, GLUE, and ARC. However, notable performance disparities emerge between text-based input and synthesized image input on mathematical reasoning tasks (e.g., MMLU-Pro, GSM8K) and programming tasks (e.g., MBPP). Phi-3.5-Vision exhibits consistently poor performance across all vision tasks, primarily due to its insufficient instruction-following capabilities.}
    \label{fig:performance_comparison}
\end{figure*}
\label{sec:experiment}
\noindent \textbf{Baseline} We establish the baseline by using the same VLMs with text-only prompts. To ensure fairness, we employ identical prompts and add the instruction \textit{“Please follow the instruction in the image”} when applying \method. This ensures that the VLMs can correctly process instructions embedded within images. Ideally, the baseline and \method should yield equivalent performance. This comparison helps identify areas for improvement in existing VLMs.

\noindent \textbf{Metrics} For QA tasks (\textit{WikiSS-QA}, \textit{SlidesVQA}, \textit{TableBench}), we use \textit{ROUGE-L}, which measures the longest common subsequence between prediction and reference to approximate answer overlap. We choose it for convenience and comparability, and expect other semantic metrics (e.g., BERTScore, LLM judges) to show similar trends. For classification benchmarks, including \textit{MMLU-Pro}, \textit{GLUE}, \textit{SuperGLUE}, \textit{ARC}, and \textit{MathVerse}, we use accuracy, which directly reflects the model's performance in selecting correct options. For \textit{GLUE} and \textit{SuperGLUE}, we follow their standard evaluation protocols, utilizing task-specific metrics such as Matthews correlation, F1 score, and Pearson correlation. For the code generation task \textit{MBPP}, we evaluate performance using the pass@1 rate, which measures whether the generated code successfully passes all test cases. For the mathematical reasoning dataset \textit{GSM8K}, we employ exact match accuracy, as these problems require precise numerical answers. For the visualization subtask of \textit{TableBench}, following the original codebase, we treat it as a code generation task and evaluate the correctness of the generated visualizations.

\begin{figure*}[t]
    \centering
    \includegraphics[width=\textwidth]{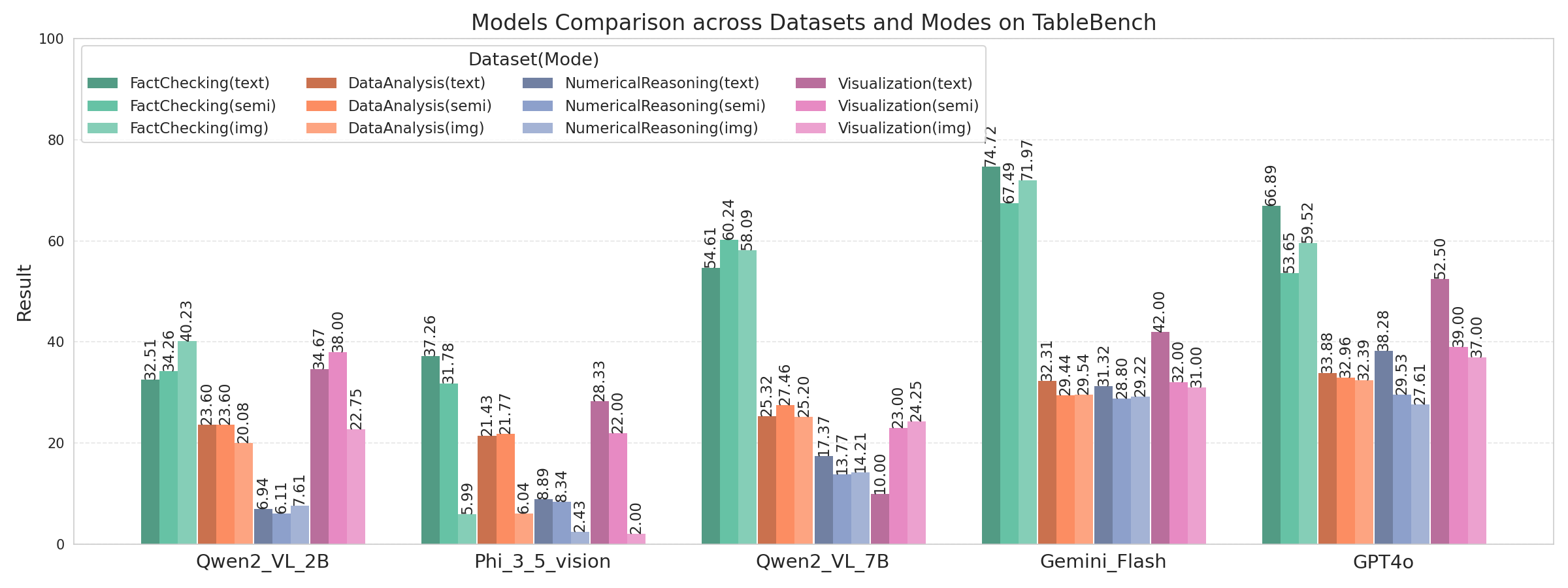} 
    \caption{The performance of the \textbf{structured} dataset. We report all the subsets for the TableBench. In the \textit{semi} setting, questions were presented as text, while tables were rendered as synthetic images. We observed that for tasks involving reasoning (numerical reasoning) and coding (visualization subset), synthetic images yielded inferior performance compared to text. However, for tasks emphasizing semantic understanding, such as data analysis and fact checking, synthetic images achieved performance comparable to or even surpassing text. Additionally, we found that the semi approach often performed worse than either text or synthetic images individually, providing insights into potential limitations and future directions for leveraging vision-language models (VLMs).}
    \label{fig:performance_comparison_structure}
\end{figure*}

\noindent\textbf{Model Selection} To validate \textsc{PixelWorld}, we selected a diverse set of vision-language models (VLMs) with varying scales to ensure the robustness and generalizability of our findings. It also allowed us to analyze the behavior of models across different sizes. We evaluated several widely used vision-language models (VLMs), including Qwen2VL-2B \citep{Qwen2VL}, Phi-3.5-3.2B \citep{abdin2024phi3technicalreporthighly}, Qwen2VL-7B \citep{Qwen2VL}, Gemini-Flash \citep{gemini}\footnote{We use \texttt{gemini-1.5-flash-002}, which was the latest available version during this study.}, and GPT-4o \citep{openai_gpt4o}.

\subsection{Text Input}
\label{sec:experiment_text}
Figure \ref{fig:performance_comparison} reports model accuracy on text-only datasets (e.g., ARC, MMLU-Pro, GLUE, GSM8K, SuperGLUE, MBPP). Two major insights emerge:

\noindent \textbf{Better Transferability in Larger Models} Larger language models (e.g., GPT-4o, Gemini-Flash) exhibit better transferability between text and image-based performance, while smaller models struggle with both transferability and instruction following. For instance, on the ARC dataset, GPT-4o's performance declines by only 0.59 points when transitioning from text to synthetic images, whereas the smaller Qwen2-VL-2B suffers a substantial 21.73-point drop (from approximately 68.61 to 46.88). This trend suggests that more capable models preserve their reasoning abilities across modalities, while smaller models face greater difficulty. Additionally, smaller models (e.g., Phi-3.5-vision) not only show weaker overall performance on standard benchmarks but also struggle significantly when instructions are presented as images. Their performance consistently lags behind that of larger models, particularly on tasks like MBPP. This supports \textit{Insight 3} in Figure \ref{fig:peap}.

\noindent \textbf{Performance Degradation with More Complex Tasks} We observe significant drops on benchmarks requiring advanced reasoning, such as mathematical, coding or domain-specific tasks. For example, when moving from text to image inputs on the MMLU-Pro dataset, GPT-4o exhibits a drop of more than 25 points. In contrast, on GLUE and SuperGLUE, the decline remains under 5 points. These findings indicate that while existing large models achieve comparable performance between text and visual modalities on simpler tasks, a gap still exists at a deeper level in visual-based and text-based understanding, demonstrating room for improvement in modality adaptation training.

\subsection{Structured Input} \label{sec:experiment_structure} 
Figure~\ref{fig:performance_comparison_structure} summarizes model performance on four TableBench subsets: Fact Checking, Data Analysis, Numerical Reasoning, and Visualization.

\noindent \textbf{Reasoning Complexity Impacts Performance} Fact Checking and Data Analysis show moderate performance drops, as they rely on semantic understanding. In contrast, Numerical Reasoning and Visualization—requiring more intricate reasoning and coding—exhibit larger declines when switching to synthetic images. Combined with \textit{``Performance Degradation with More Complex Tasks''} in Section~\ref{sec:experiment_text}, this supports \textit{Insight 2} in Figure \ref{fig:peap}.

\noindent \textbf{Smaller Performance Gaps with Structured Data} Compared to text-only tasks, structured tasks show smaller performance gaps between text and image inputs. Notably, Qwen2VL-2B even outperforms its text-based results on Fact Checking, suggesting robust visual representations can aid semantic tasks in smaller models.

\noindent \textbf{Challenges with Mixed-Modality Inputs} The ``semi'' format—where tables appear as images while questions remain text-based—performs worse than either fully text-based or fully image-based formats. This suggests that conventional VQA approaches, which process text and images using separate encoders, may be more susceptible to performance bottlenecks. As multimodal scenarios become increasingly prevalent, \method is expected to demonstrate superior performance compared to mixed-modality methods.
\begin{figure}[t]
    \centering
\includegraphics[width=0.5\columnwidth]{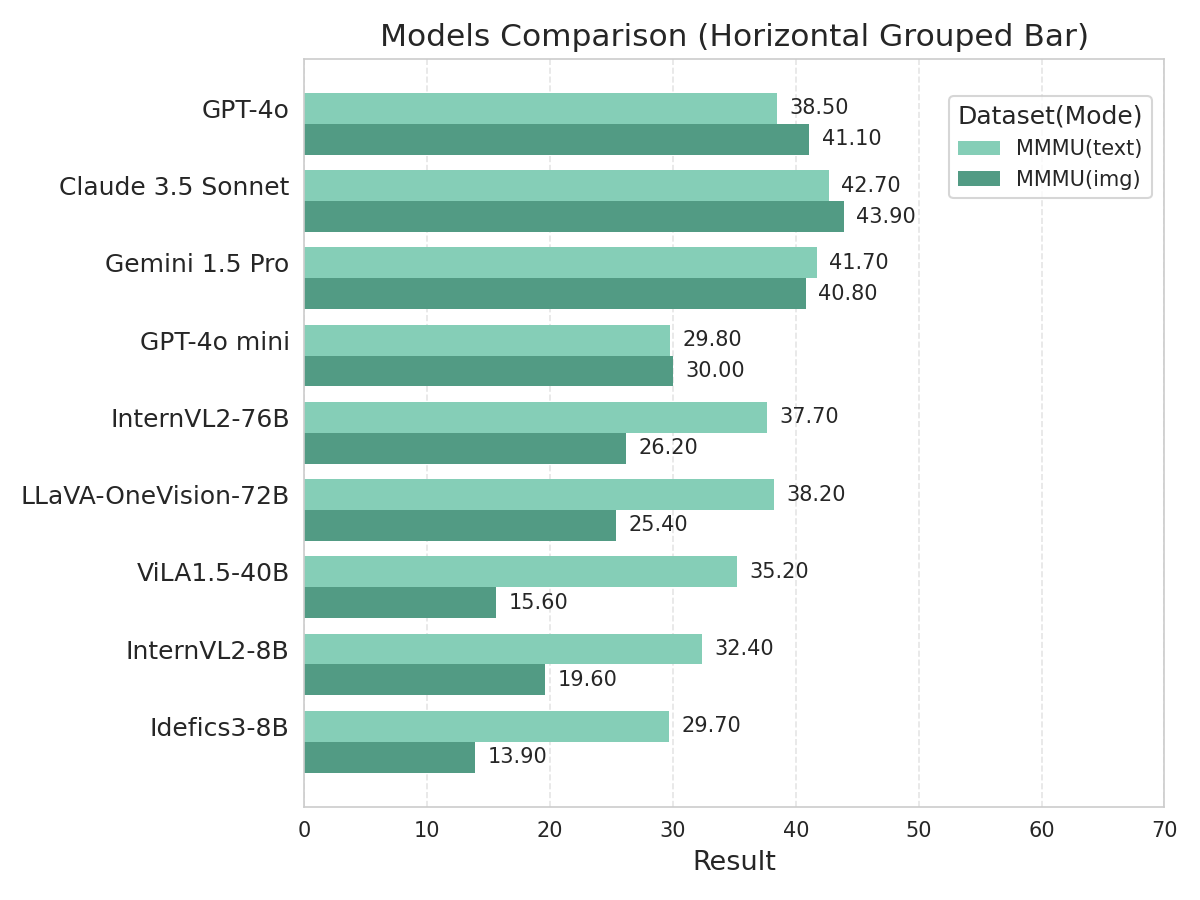} 
    \caption{The performance of the \textbf{multimodal} dataset (MMMU-Pro). We adopt the result reported by the origin paper. We can observe that strong models perform better in \method.}
    \label{fig:performance_comparison_mmmu}
\end{figure}

\begin{figure*}[t]
    \centering
    \includegraphics[width=\textwidth]{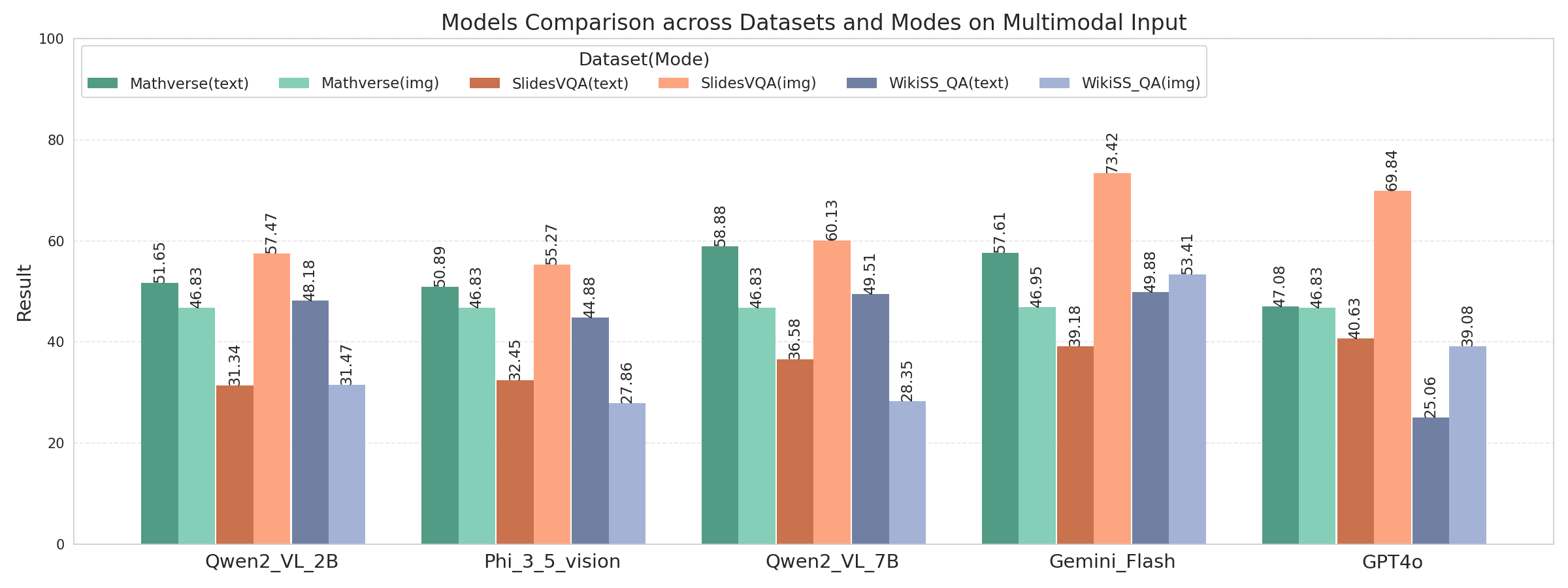} 
    \caption{The performance of the \textbf{multimodal} datasets (except MMMU-Pro). We compare text-only and vision-only subsets in Mathverse, while SlidesVQA and WikiSS-QA are evaluated as VQA tasks. Larger models perform better on text-based tasks with more modalities. GPT-4o tends to generate longer responses in long-context QA, leading to performance degradation on WikiSS-QA.}
    \label{fig:performance_comparison_multi}
\end{figure*}

\subsection{Multimodal Input}
\label{sec:experiment_multi}
Figure~\ref{fig:performance_comparison_multi} presents model performance on multimodal datasets, including text-only and vision-only subsets of Mathverse and VQA tasks like SlidesVQA and WikiSS-QA. Results on MMMU-Pro (Figure~\ref{fig:performance_comparison_mmmu}) use reported values from the original paper. Three key observations emerge:

\noindent \textbf{Image Inputs Enhance Disambiguation} Incorporating images improves performance by reducing ambiguity compared to text-only benchmarks. In SlidesVQA, all models outperform their text-only baselines, while in WikiSS-QA and MMLU-Pro, visual context provides clarifying information, leading to accuracy gains in larger models. Combined with \textit{``Smaller Performance Gaps with Structured Data''} in Section~\ref{sec:experiment_structure}, this supports \textit{Insight 1} in Figure \ref{fig:peap}.

\noindent \textbf{Challenges in Complex Reasoning} While multimodal inputs aid basic tasks, complex reasoning remains a bottleneck. In Mathverse, visual cues help but fail to support multi-step logical deductions. Even Gemini-Flash shows accuracy drops on intricate reasoning tasks. Additionally, WikiSS-QA poses challenges due to its long-context nature. Smaller models struggle with \method, and GPT-4o underperforms in token-based tasks, highlighting difficulties in processing extended contextual dependencies. This aligns with Sections~\ref{sec:experiment_text} and~\ref{sec:experiment_structure}.

\noindent \textbf{Larger Models Benefit More from Multimodal Data} Larger models gain more from multimodal inputs. On SlidesVQA, Gemini\_Flash improves by 34.24 points, compared to Qwen2-VL-7B’s 23.55-point boost. This suggests that larger models, with more extensive prior knowledge and advanced architectures, leverage multimodal data more effectively than smaller models. %
 
\begin{figure*}[t]
    \centering
    \makebox[0.9\textwidth][l]{
        \hspace{-1.05cm} 
        \includegraphics[width=\textwidth]{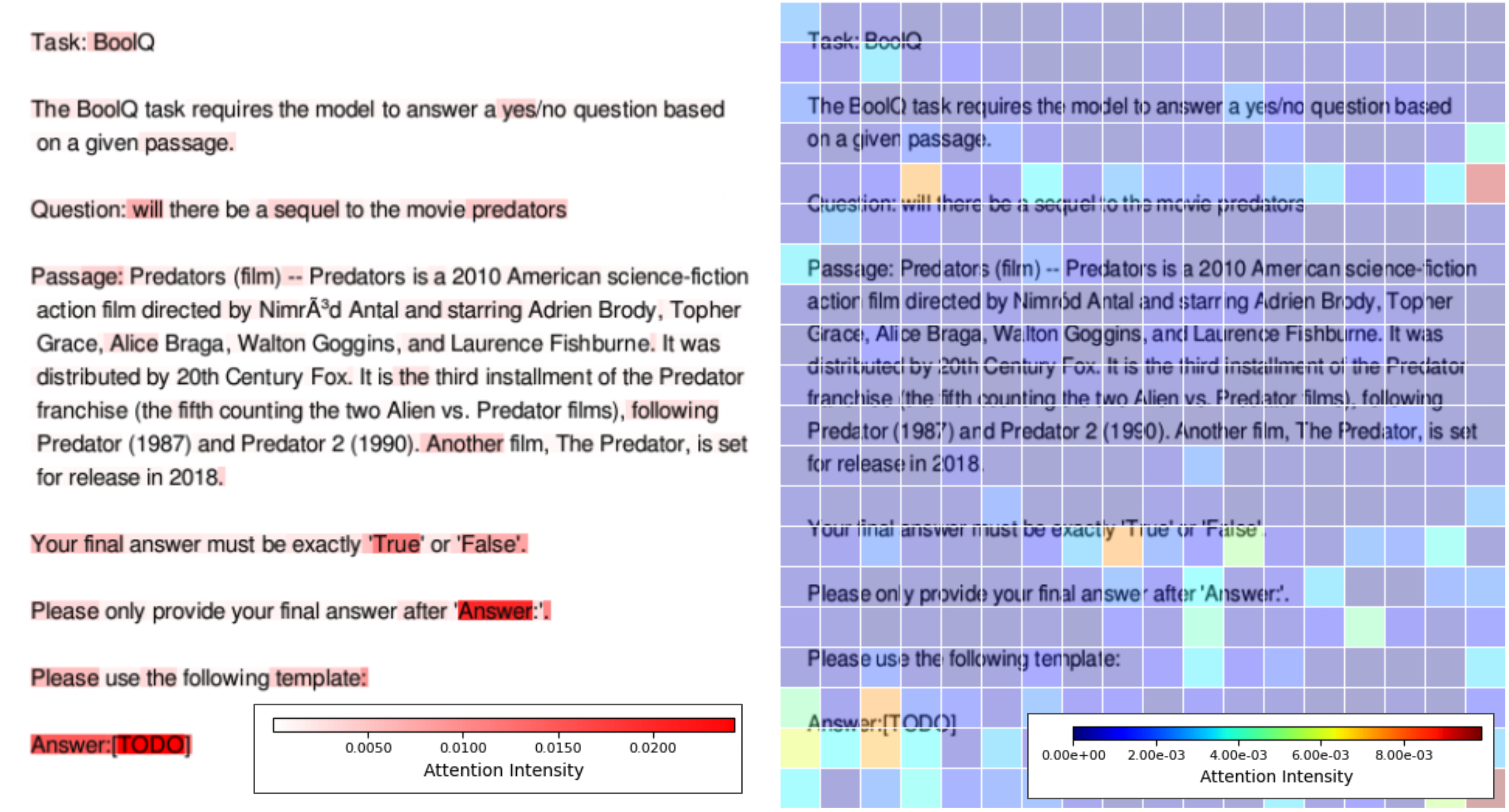} 
    }
    \caption{Last Layer Attention Heatmap on Qwen2VL-7B between token-based (left) and pixel-based (right) inference. Although the overall attention intensity on image inputs is generally lower, both modalities exhibit highly similar attention patterns.}
    \label{fig:heatmap}
\end{figure*}

\section{Discussion}
\label{sec:dis}

\subsection{Q1: Does \method have the same attention?}
\label{sec:dis_1}

To examine whether VLMs attend to similar regions when processing textual and image inputs, we visualize the average attention map of the final layer in Qwen2-VL-7B using a heatmap (Figure~\ref{fig:heatmap}). Specifically, we analyze the model’s behavior on a \textit{BoolQ} example from SuperGLUE, comparing its attention patterns under text-based and image-based inference. Similar attention behaviors are observed across different datasets; more examples are shown in Appendix~\ref{app:heatmaps}.

\noindent\textbf{Computation of Heatmaps.}  
The attention heatmaps are computed during greedy decoding by averaging the last-layer attention across tokens. For multi-head attention, we apply a simple mean across heads. Formally, the attention weight for position $i$ is given by:

$$
\text{Heatmap}(i) \;=\; \frac{1}{H} \sum_{h=1}^{H} \big| A^{(L)}_{h}[t, i] \big|, 
\quad i \in [s, e)
$$

where $H$ denotes the number of attention heads, $L$ is the index of the last layer, $A^{(L)}_{h}[t, i]$ represents the attention weight from token $t$ to token $i$ in head $h$, and $[s, e)$ corresponds to the decoding range during greedy decoding.

\noindent\textbf{Observations.}  
As shown in Figure~\ref{fig:heatmap}, Qwen2-VL-7B consistently focuses on task-relevant elements such as the question prompt (``will there be a sequel ...''), salient passage keywords (e.g., ``film'', ``starring'', ``Alice''), and the expected answer format (``Answer: True/False''). This pattern remains stable across both textual and visual representations, suggesting that the model exhibits largely comparable attention behavior regardless of input modality. However, we also observe that certain blank patches in the image-based inputs occasionally receive disproportionately high attention weights, indicating that while the visual encoder aligns closely with the text encoder, it still introduces redundant activations.

\subsection{Q2: How to make PEAP more efficient?}
\label{sec:dis_2}
\begin{table}[t]
    \centering \small
    \setlength\tabcolsep{4pt}
    \begin{tabular}{lccc}
\toprule
& \multicolumn{3}{c}{SuperGLUE Evaluation Results} \\
\cmidrule(lr){2-4}
\textbf{Task}       & \textbf{Text} & \textbf{\method} & \textbf{\method-Fast} \\
\midrule
BoolQ               & 79.69\%           & 82.11\%                      & 80.89\%                      \\
CB                  & 67.70\%           & 40.77\%                      & 39.57\%                      \\
COPA                & 93.00\%           & 91.00\%                      & 86.00\%                      \\
MultiRC             & 65.90\%           & 61.28\%                      & 60.80\%                      \\
ReCoRD              & 12.54\%           & 5.94\%                       & 6.08\%                       \\
RTE                 & 82.31\%           & 72.92\%                      & 77.26\%                      \\
WiC                 & 53.29\%           & 55.80\%                      & 55.64\%                      \\
WSC                 & 63.46\%           & 65.38\%                      & 59.62\%                      \\
\midrule
\textbf{Final Score} & 64.74\%           & 59.40\%                      & 58.23\%                      \\
\bottomrule
    \end{tabular}
    \caption{Performance of \textit{Qwen2VL-7B} on SuperGLUE dataset by  Text, \method and \method-Fast. We can observe the comparable performance between \method and \method-Fast.}
    \label{tab:superglue_results}
\end{table}

\begin{table}[t]
    \centering \small
    \hspace*{-0.6cm}
    \setlength\tabcolsep{4pt}
    \begin{tabular}{lcccccc}
\toprule
& \multicolumn{3}{c}{Inference Time (s)} & \multicolumn{2}{c}{Overhead (\%)} \\
\cmidrule(lr){2-4} \cmidrule(lr){5-6}
\textbf{Subset} & \textbf{Text} & \textbf{\method} & \textbf{\method-Fast} & \textbf{\method} & \textbf{\method-Fast} \\
\midrule
BoolQ           & 369           & 1,381               & 906           & 274.80          & 145.55          \\
CB              & 8             & 22                  & 15            & 175.00          & 87.50           \\
COPA            & 39            & 38                  & 22            & -2.56           & -43.59          \\
MultiRC         & 609           & 3,861               & 2,550         & 534.80          & 318.71          \\
ReCoRD          & 7,016         & 19,012              & 14,288        & 171.01          & 103.72          \\
RTE             & 68            & 117                 & 92            & 72.06           & 35.29           \\
WiC             & 69            & 224                 & 157           & 224.64          & 127.54          \\
WSC             & 11            & 36                  & 27            & 227.27          & 145.45          \\
\midrule
\textbf{Total}  & 8,089         & 24,690              & 18,051        & 205.27          & 123.19          \\
\bottomrule
    \end{tabular}
    \caption{Inference Time (s) of \textit{Qwen2VL-7B} on SuperGLUE dataset with single A100 server by \method and \method-Fast. We can observe a 82.08\% overhead reduce on \method-Fast method. Overhead is calculated as the percentage increase in time relative to the text method.}
    \label{tab:inference_time}
\end{table}

As a trade-off for generalization, image-based inference often requires significantly more computational resources than text-based inference. This is partly due to the additional overhead from the ViT backbone and higher redundancy in image tokens. To estimate the performance gap quantitatively, we conducted experiments on SuperGLUE ({Table~\ref{tab:superglue_results}}). The results show that inference latency for image-based inputs can exceed text-based methods by 150\% to 250\%. 

To reduce redundancy in visual inputs, we propose \textbf{\method-Fast}, which first identifies empty patches via a simple variance-based threshold—if the pixel-value variance in a patch is lower than a preset threshold, that patch is treated as empty and is pruned from all attention computations. Crucially, we preserve the original positional embeddings for the remaining tokens, ensuring no loss of spatial layout perception. This strategy aligns with how humans naturally focus on salient regions rather than blank spaces, thereby significantly reducing context length without sacrificing structural information. Testing \method-Fast on SuperGLUE reveals a minor accuracy drop of only 1.17\% (Table~\ref{tab:superglue_results}). More importantly, the average overhead decreases from 205.27\% to 123.19\%, yielding an 82.98\% reduction (Table~\ref{tab:inference_time}). These results demonstrate that removing empty patches offers substantial computational savings while maintaining strong performance, making image-based inference more practical for real-world deployments. Attention heatmap between \method and \method-Fast are shown in Appendix \ref{sec:heatmap_compare_fast}.

\subsection{Q3: Is \method sensitive to the prompting method?}
\label{sec:dis_3}
\begin{table*}[t]
    \centering \small
    \setlength\tabcolsep{6pt}
    \begin{tabular}{lcccccc}
\toprule
\textbf{Metric} & \multicolumn{2}{c}{\textbf{Direct}} & \multicolumn{2}{c}{\textbf{CoT}} & \multicolumn{2}{c}{\textbf{Improve (CoT - Direct)}} \\
\cmidrule(lr){2-3} \cmidrule(lr){4-5} \cmidrule(lr){6-7}
& \textbf{Text} & \textbf{\method} & \textbf{Text} & \textbf{\method} & \textbf{Text} & \textbf{\method} \\
\midrule
{BoolQ}   & 79.88\% & 81.71\% & 81.13\% & 80.73\% & 1.25\%  & -0.98\% \\
{CB}      & 67.70\% & 34.78\% & 81.04\% & 59.57\% & 13.34\% & 24.79\% \\
{COPA}    & 93.00\% & 87.00\% & 89.00\% & 83.00\% & -4.00\% & -4.00\% \\
{MultiRC} & 65.73\% & 62.28\% & 69.08\% & 60.41\% & 3.35\%  & -1.87\% \\
{ReCoRD}  & 12.50\% & 5.88\%  & 6.37\%  & 4.66\%  & -6.13\% & -1.22\% \\
{RTE}     & 82.31\% & 72.92\% & 83.03\% & 77.26\% & 0.72\%  & 4.34\%  \\
{WiC}     & 52.82\% & 54.39\% & 54.39\% & 53.92\% & 1.57\%  & -0.47\% \\
{WSC}     & 65.38\% & 61.54\% & 57.69\% & 61.54\% & -7.69\% & 0.00\%  \\
\midrule
{Overall} & 64.92\% & 57.56\% & 65.22\% & 60.14\% & 0.30\%  & 2.58\%  \\
\bottomrule
    \end{tabular}
    \caption{Comparison of Direct and CoT performance across Text and Image modalities, along with their respective improvements (CoT - Direct), presented as percentages.}
    \label{tab:base_cot_improve}
\end{table*}

In Section \ref{sec:experiment}, results on purely textual synthesis tasks show that image-based inputs consistently underperform text inputs, likely due to dataset domain gaps and weaker instruction following on image representations. To address this, we applied CoT-style prompts to the SuperGLUE dataset to enhance cross-domain instruction following (Table \ref{tab:base_cot_improve}). Notably, Qwen2VL-7B showed significant improvements in tasks where image input underperformed compared to text input, such as CB and RTE. Overall, CoT prompts improved image input performance by 2.58\%, surpassing the 0.3\% improvement observed for text input.

\section{Related Work}
\paragraph{Multimodal Large Language Models and Benchmarks} 
Recent progress in multimodal AI has led to the development of models like GPT-4o \citep{openai_gpt4o}, Gemini \citep{gemini}, and Claude-3.5 \citep{anthropic_claude3.5}, which integrate vision-based training to improve instruction-following capabilities. Benchmarks for these models have evolved from task-specific datasets, such as VQA \citep{agrawal2016vqavisualquestionanswering} and DocVQA \citep{mathew2021docvqadatasetvqadocument}, to more comprehensive evaluations, including MMMU-Pro \citep{yue2024mmmuprorobustmultidisciplinemultimodal}, MMBench \citep{liu2024mmbenchmultimodalmodelallaround}, and MegaBench \citep{megabench}. However, most current research focuses on the semantic understanding of visual content, with only a few benchmarks—such as MathVerse \citep{zhang2025mathverse} and MMMU-Pro \citep{yue2024mmmuprorobustmultidisciplinemultimodal}—addressing text recognition and comprehension within images. Our work shifts the focus towards evaluating how well large language models understand language through visual input compared to traditional token-based input.

\paragraph{Screenshot LMs} Recent studies have demonstrated that pretraining on synthetic screenshots can enable vision-language models (VLMs) to achieve performance comparable to that of BERT on language modeling tasks \citep{https://doi.org/10.48550/arxiv.2210.03347, rust-etal-2023-pixel, gao2024improvinglanguageunderstandingscreenshots}. This approach allows models to better capture text structures without relying on OCR-based methods. Furthermore, our analysis highlights a performance gap between existing VLMs on vision-based tasks and their text-only counterparts, particularly in the absence of relevant pretraining. Interestingly, in certain scenarios, VLMs perform as well as or even better than text-only models, underscoring the potential of this research direction. In the context of document retrieval, recent advancements \citep{faysse2024colpaliefficientdocumentretrieval, ma-etal-2024-unifying} have shown that large-scale pretraining on screenshots can outperform traditional OCR-based methods, further reinforcing the advantages of vision-language pretraining.

\paragraph{Language Tokenization} Tokenization methods, such as Byte Pair Encoding (BPE) \citep{bpe,nmt}, are widely used in language modeling, but recent studies suggest that they may not always be optimal. For instance, MegaByte \citep{yu2023megabyte} demonstrated that fixed-length tokenization can improve both computational efficiency and cross-modal capabilities. Similarly, BLT \citep{pagnoni2024byte} proposed entropy-based tokenization, while LCM \citep{lcmteam2024largeconceptmodelslanguage} emphasized the benefits of processing higher-level semantic concepts rather than individual tokens. Inspired by these approaches, we explore whether adaptive image patches can effectively infer textual meaning. At a higher level, we investigate the unification of text and image inputs into a shared representation space, enabling reasoning through abstract semantic concepts rather than traditional token-based methods.

\section{Conclusion}

We present \textsc{PixelWorld}, a benchmark that renders text, tables, code, and images as pixels, enabling direct evaluation of the \emph{Perceive Everything as Pixels} (\method) paradigm. Experiments yield three main takeaways.  
\textbf{(1) Semantic understanding:} \method\ performs on par with token-based baselines on sentence- and paragraph-level tasks, and its patch-level attention exhibits similar global structures to token attention, suggesting partial transfer of language modeling behavior into the visual domain.  
\textbf{(2) Reasoning:} Performance drops on math, logic, and program-repair benchmarks, though Chain-of-Thought prompting mitigates but does not close this gap, highlighting the continued need for explicit reasoning structure.  
\textbf{(3) Multimodal perception:} Pixel-based inputs outperform OCR pipelines on websites, slides, and documents by preserving spatial context and avoiding recognition noise.  
To alleviate the higher computational cost of pixel inputs, we propose \method-Fast, which prunes blank patches and achieves up to a \(3\times\) inference speedup with minimal accuracy loss. Overall, these findings illustrate both the potential and limitations of \method, positioning \textsc{PixelWorld} as a diagnostic and reproducible benchmark for studying unified vision--language representations and guiding future work on efficiency and reasoning in multimodal agents.

\bibliography{main}
\bibliographystyle{tmlr}

\appendix
\section{Example Input}
\label{sec:example_input}
\begin{figure*}[t]
    \centering
    \includegraphics[width=0.8\textwidth]{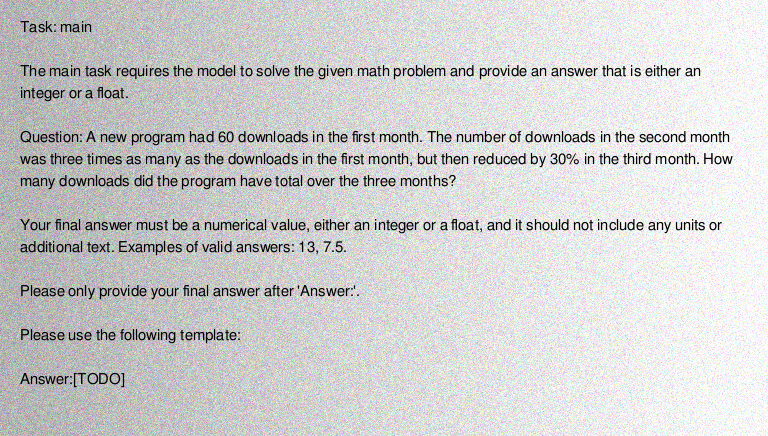} 
    \caption{An example input of GSM8K dataset, using Direct Prompt.}
    \label{fig:example_input}
\end{figure*}

\begin{figure*}[t]
    \centering
    \includegraphics[width=0.8\textwidth]{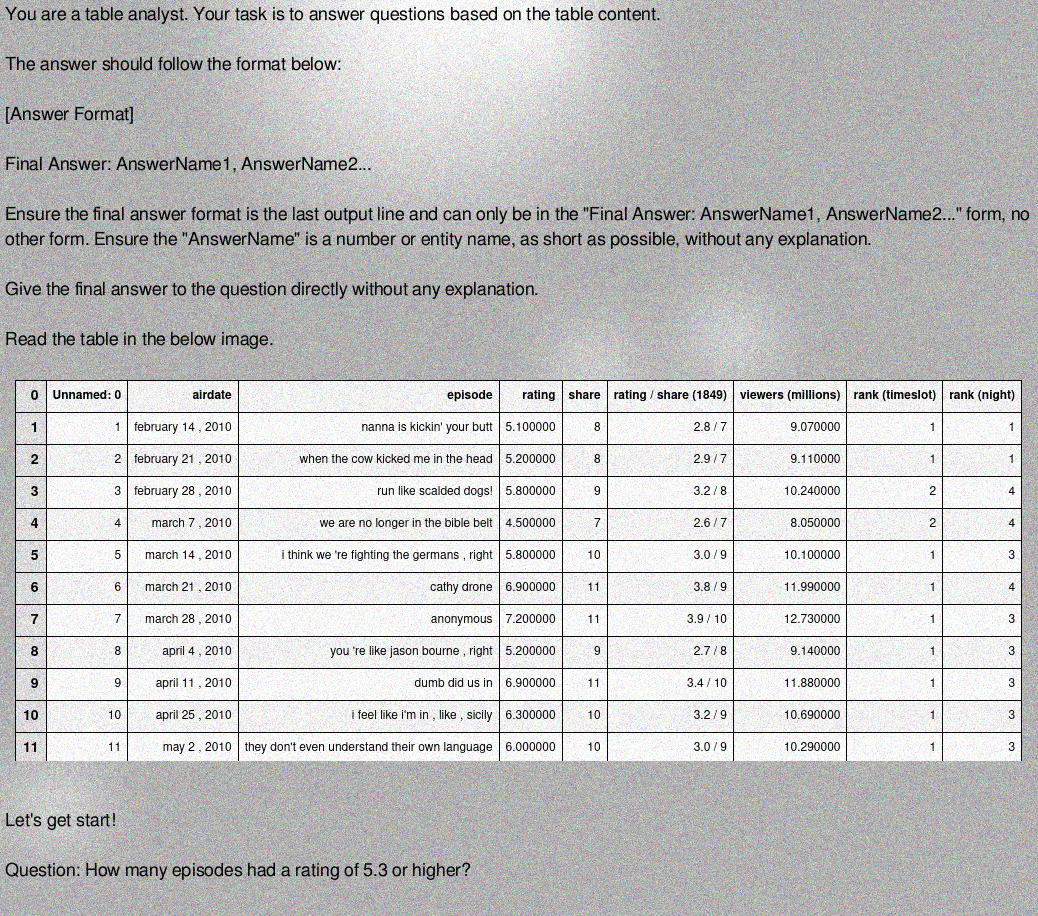} 
    \caption{An example input of TableBench dataset, using Direct Prompt.}
    \label{fig:example_input2}
\end{figure*}
Figure \ref{fig:example_input} and Figure \ref{fig:example_input2} gives two examples about the vision input.

\section{Attention Heatmap before and after ImageFast Method}
\label{sec:heatmap_compare_fast}
\begin{figure*}[t]
    \centering
    \makebox[\textwidth][l]{
        \includegraphics[width=1\textwidth]{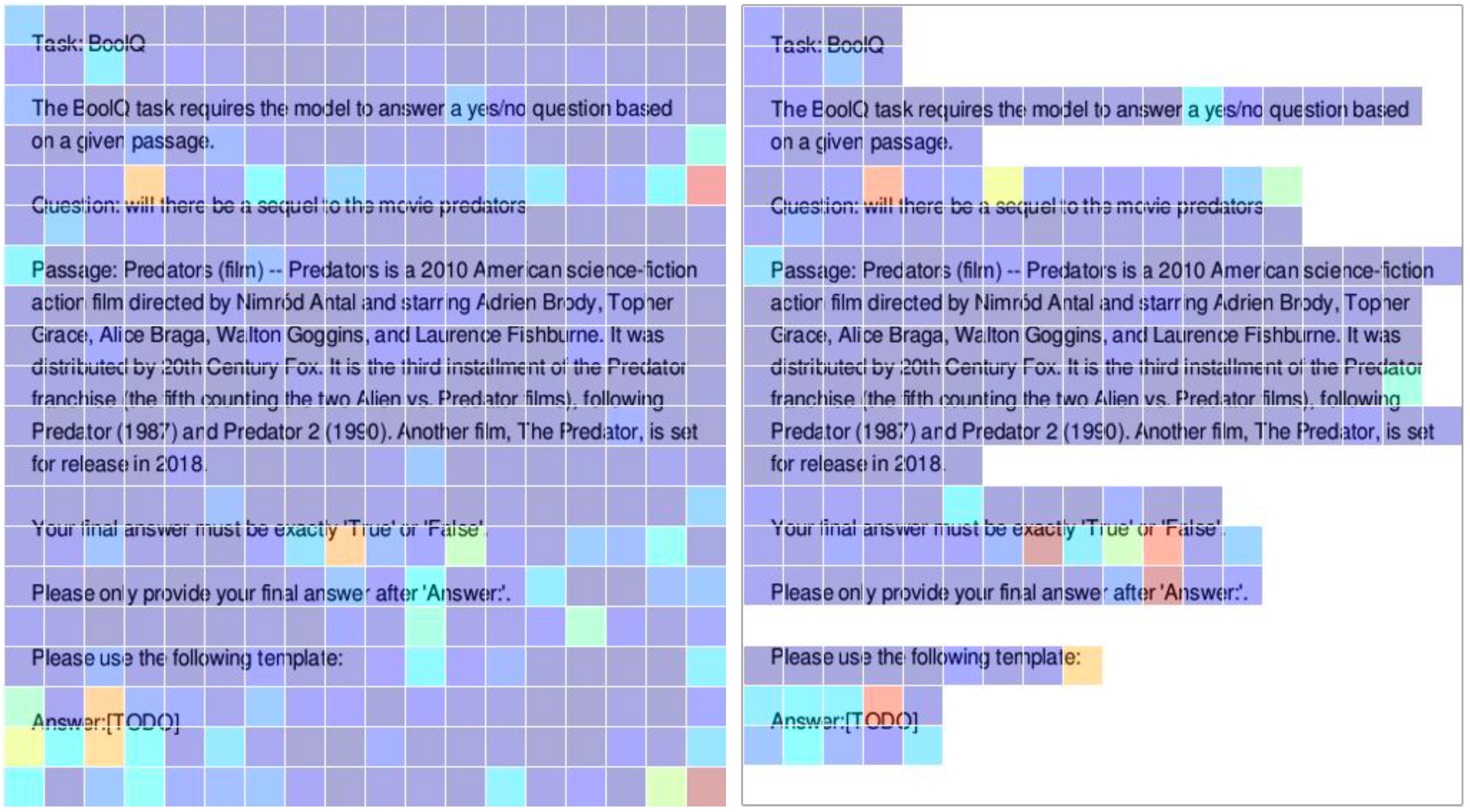} 
    }
    \caption{Last Layer Attention Heatmap on Qwen2VL-7B between \method (left) and \method-Fast (right).}
    \label{fig:heatmap_Fast}
\end{figure*}

Figure \ref{fig:heatmap_Fast} presents a heatmap comparison between \method and \method-Fast. \method-Fast effectively reduces redundant patches while preserving attention on key regions.

\section{Attention Heatmap Comparison Between Datasets}
\label{app:heatmaps}
We provide additional attention visualizations on three representative datasets—\textit{MBPP}, \textit{MMLU-Pro}, and \textit{MathVerse}—to illustrate how attention patterns vary across program synthesis, reasoning, and STEM tasks.

\begin{figure*}[t]
    \centering
    \makebox[0.9\textwidth][l]{
        \hspace{-1.05cm} 
        \includegraphics[width=\textwidth]{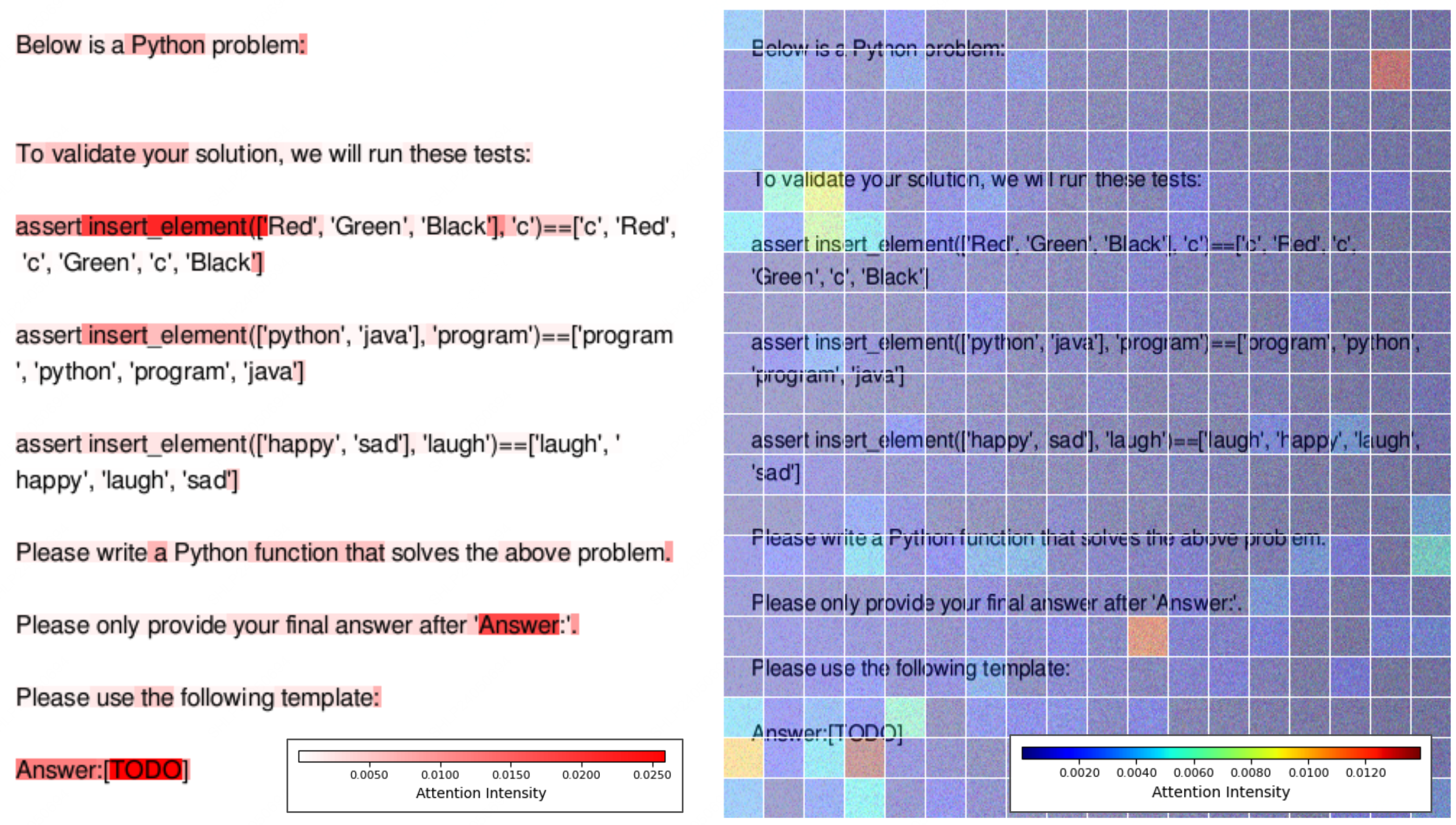} 
    }
    \caption{Example 1: Last Layer Attention Heatmap on Qwen2VL-7B from MBPP.}
    \label{fig:heatmap_MBPP}
\end{figure*}

\begin{figure*}[t]
    \centering
    \makebox[0.9\textwidth][l]{
        \hspace{-1.05cm} 
        \includegraphics[width=\textwidth]{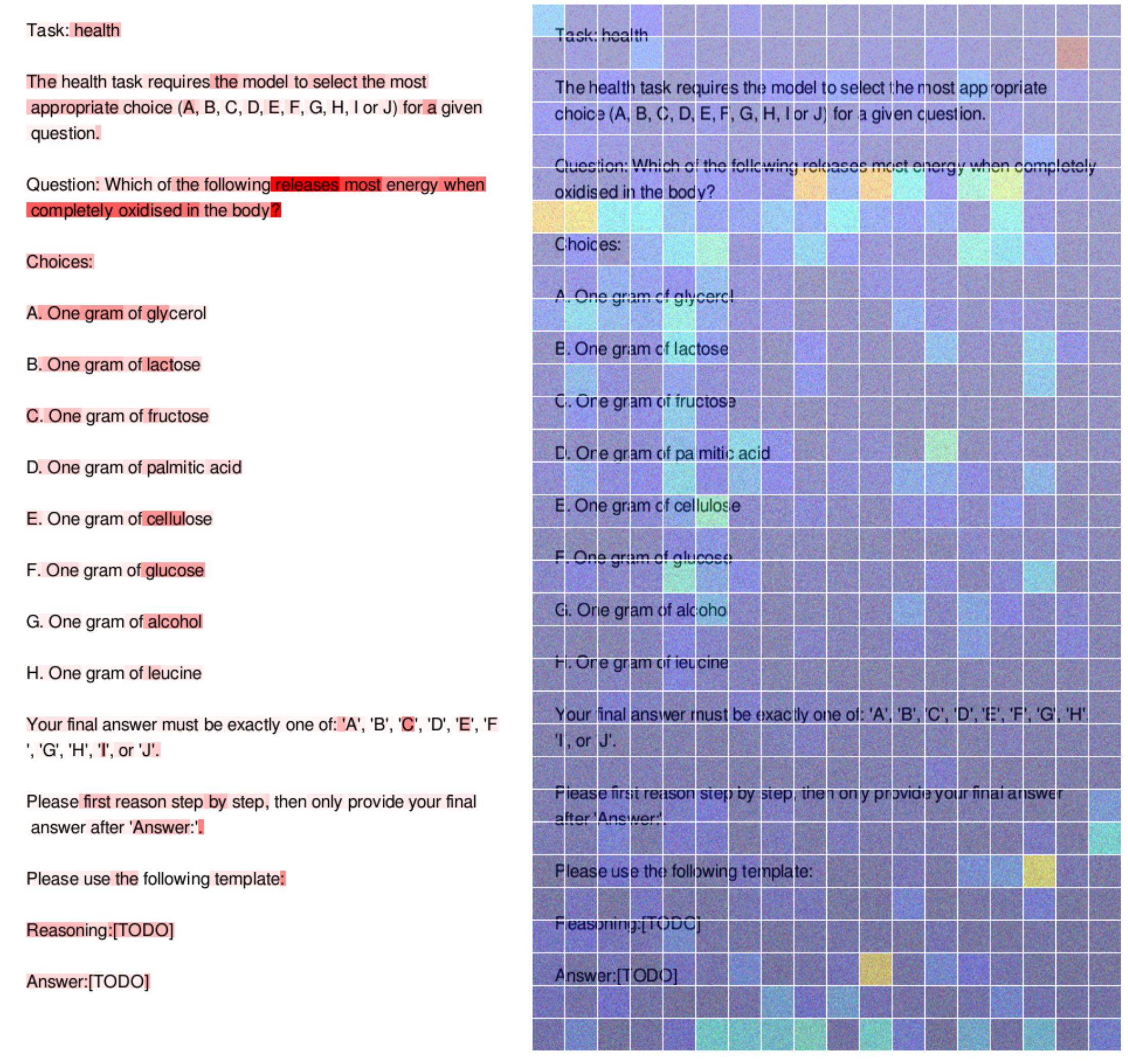} 
    }
    \caption{Example 2: Last Layer Attention Heatmap on Qwen2VL-7B from MMLU-Pro.}
    \label{fig:heatmap_MMLU_Pro}
\end{figure*}
\begin{figure*}[t]
    \centering
    \makebox[0.9\textwidth][l]{
        \hspace{-1.05cm} 
        \includegraphics[width=\textwidth]{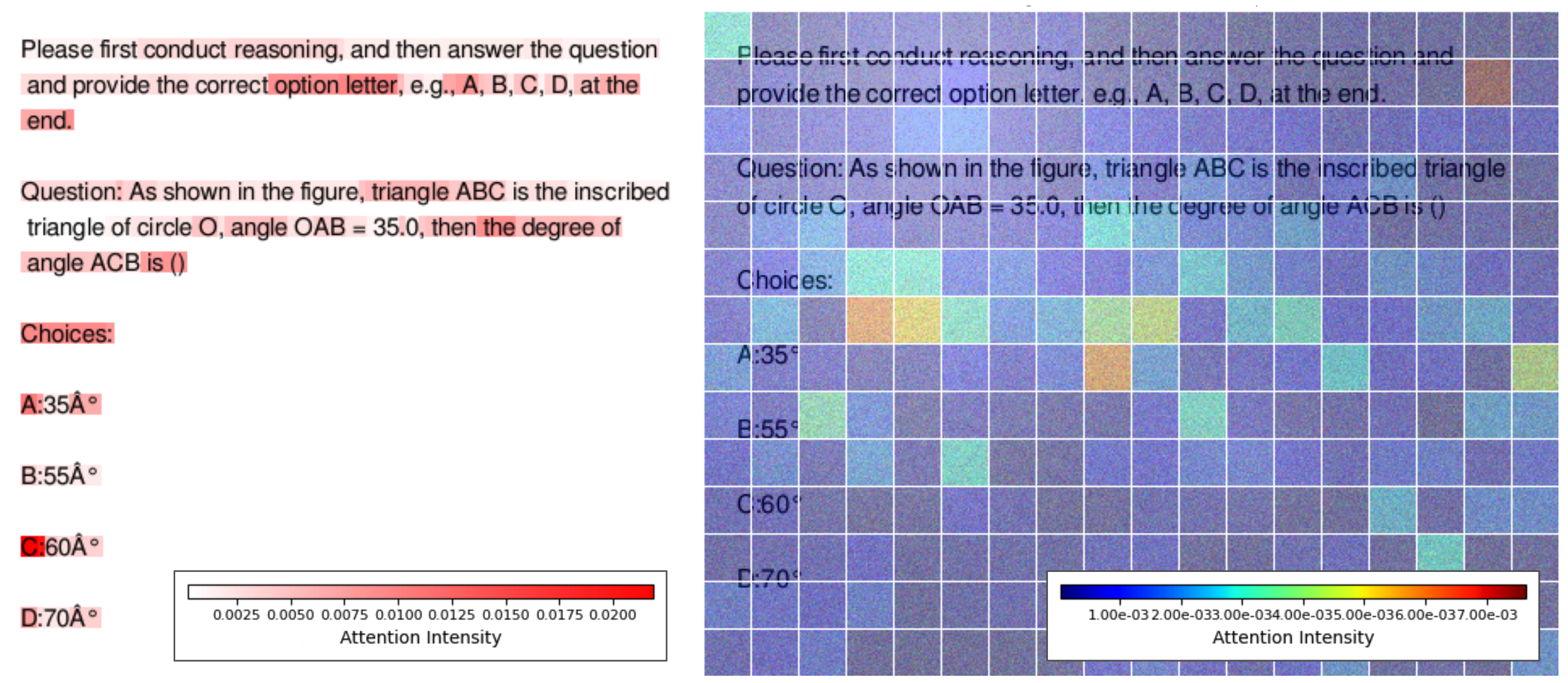} 
    }
    \caption{Example 3: Last Layer Attention Heatmap on Qwen2VL-7B from MathVerse.}
    \label{fig:heatmap_Mathverse}
\end{figure*}

\section{Broader Impact Statement}

This work explores a unified pixel-based perception paradigm that eliminates the need for separate text and image tokenization. While such unification can simplify multimodal pipelines and reduce reliance on OCR systems, it also introduces new risks. The computational cost of pixel-based models remains significantly higher than text-based counterparts, which may limit accessibility and increase the carbon footprint of large-scale training and deployment. Furthermore, because pixel inputs may contain sensitive visual information, researchers must ensure that data synthesis and collection comply with privacy and ethical standards.

On the positive side, the PixelWorld benchmark provides a transparent and reproducible foundation for assessing multimodal understanding, encouraging fair comparisons across models and modalities. By highlighting where pixel-based representations succeed and fail, this work aims to guide the community toward more efficient and interpretable multimodal systems, fostering broader exploration of unified perception without compromising ethical or environmental considerations.
\end{document}

%% file: math_commands.tex
\usepackage{amsmath,amsfonts,bm}

\def\eqref#1{equation~\ref{#1}}

\def\1{\bm{1}}

\DeclareMathAlphabet{\mathsfit}{\encodingdefault}{\sfdefault}{m}{sl}
\SetMathAlphabet{\mathsfit}{bold}{\encodingdefault}{\sfdefault}{bx}{n}
\newcommand{\tens}[1]{\bm{\mathsfit{#1}}}

\newcommand{\etens}[1]{\mathsfit{#1}}